\documentclass[letterpaper, 10 pt, conference]{ieeeconf}  

\IEEEoverridecommandlockouts                              

\usepackage{graphicx}

\let\labelindent\relax
\usepackage{kotex}
\usepackage{enumitem}
\usepackage{multirow}
\usepackage{multicol}
\usepackage{diagbox}
\usepackage{amsmath}
\usepackage[most]{tcolorbox}
\usepackage{setspace}
\usepackage{makecell}
\usepackage{color}
\usepackage{colortbl}
\usepackage{xcolor}
\usepackage{hyperref}
\usepackage{cite}

\title{\LARGE \bf
Can only LLMs do Reasoning?:\\Potential of Small Language Models in Task Planning
}

\author{Gawon Choi and Hyemin Ahn
\thanks{Gawon Choi and Hyemin Ahn are with Graduate School of Artificial Intelligence, Ulsan National Institute of Science and Technology, Ulsan, Korea. {\tt\small{ \{gawon5325, hyemin.ahn\}@unist.ac.kr}}.}
}

\begin{document}

\maketitle
\thispagestyle{empty}
\pagestyle{empty}

\setlength{\fboxrule}{0.1mm} 

\begin{abstract}
In robotics, the use of Large Language Models (LLMs) is becoming prevalent, especially for understanding human commands. In particular, LLMs are utilized as domain-agnostic task planners for high-level human commands. LLMs are capable of Chain-of-Thought (CoT) reasoning, and this allows LLMs to be task planners. However, we need to consider that modern robots still struggle to perform complex actions, and the domains where robots can be deployed are limited in practice. This leads us to pose a question: If small LMs can be trained to reason in chains within a single domain, would even small LMs be good task planners for the robots? To train smaller LMs to reason in chains, we build `COmmand-STeps datasets' (COST) consisting of high-level commands along with corresponding actionable low-level steps, via LLMs. We release not only our datasets but also the prompt templates used to generate them, to allow anyone to build datasets for their domain. We compare GPT3.5 and GPT4 with the finetuned GPT2 for task domains, in tabletop and kitchen environments, and the result shows that GPT2-medium is comparable to GPT3.5 for task planning in a specific domain. Our dataset, code, and more output samples can be found in \href{https://github.com/Gawon-Choi/small-LMs-Task-Planning}{\textcolor{blue}{https://github.com/Gawon-Choi/small-LMs-Task-Planning}}

\end{abstract}

\section{INTRODUCTION}
In recent robotics, the use of Large Language Models(LLMs) has become increasingly prevalent, particularly in understanding human instructions. Due to the huge scale of LLMs, which are commonly over 100 billion parameters, it becomes imperative to deploy LLMs on external servers, and this makes the robot highly dependent on internet speed, and any latency directly harms the efficiency.

In this paper, we examine if robots need `large-scale' language models (LMs) in practice. There are existing methodologies such as RT-2 \cite{rt2} and SayCan \cite{saycan}, and they argue the advanced capabilities of the LLMs make robots possible to accurately interpret abstract and ambiguous human commands. However, we claim that the capabilities of LLMs are too extensive compared to the actual diversity and complexity of tasks that modern robots can execute. This low task complexity is also reflected in the LLMs-based methods, which evaluate their robots in controlled environments that are engineered to look like a real office or kitchen.

We also argue that the linguistic complexity of commands for robots is quite lower than what LLMs can handle. When comparing the dataset complexity used in pretraining LLMs with those employed in training robots, the discrepancy between the low complexity of commands for robots and the advanced LLMs capability becomes evident.

\begin{figure}[t]
    \centering
    \includegraphics[width=\columnwidth]{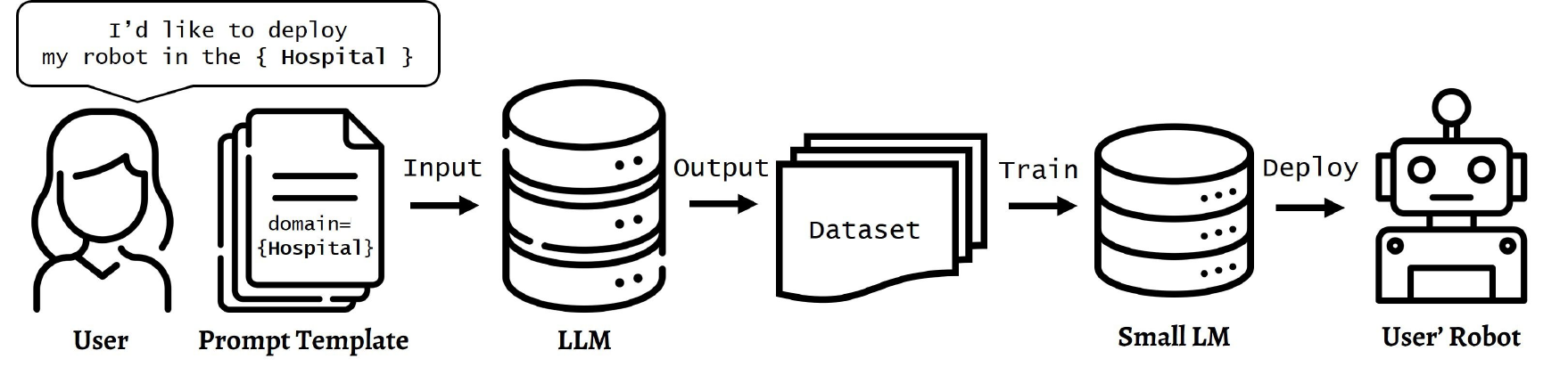}
    \vspace*{-0.7cm}
    \caption{Overview of our proposed method. When the user specifies one's domain, first, our method extract the dataset for task planning, which consists of high-level commands and low-level actionable steps, from LLMs. The dataset is used to train small LMs which will be deployed in the robot.}
    \label{fig:overall_flow}
\end{figure}
\setlength{\textfloatsep}{2pt plus 1pt minus 1pt}

One might claim that the LLMs' ability to perform Chain-of-Thought (CoT) reasoning \cite{cot_prompting} remains crucial. CoT reasoning refers to step-by-step reasoning, and it induces LLMs to describe intermediate reasoning steps to achieve the goal. Recent studies \cite{cot_good_for_bench, palm} demonstrated that designing prompts for LLMs to describe their reasoning process enhances the overall performance in the Natural Language Processing (NLP) benchmark \cite{bigbench}. LLMs can reason in chains domain-agnostic, however, we need to recall that the domains where robots can be deployed are limited in practice.

This leads us to pose a question: If small LMs can be trained to reason in chains, even within a single domain, would even smaller LMs be good task planners for the robots? To answer this question, we aim to finetune smaller LMs for CoT reasoning in a single domain. However, we face the issue of a lack of open-source datasets to train CoT reasoning. To address this issue, we utilize LLMs to build the datasets for CoT reasoning. We employ GPT3.5 \cite{instruct_gpt} to generate our `COmmand-STeps Dataset' (COST dataset) consisting of abstract commands and actionable steps that break down the command into a sequence of actions to achieve the goal, such as CoT reasoning. We finetune small LMs (i.e., GPT2) with the dataset to generate action steps for the user's command.

We release not only our dataset but also the prompt templates used to generate it. It allows anyone to build datasets for their environment and conditions, and to use them for finetuning small LMs in their domain. We also compare these finetuned small LMs with LLMs in tabletop and kitchen environments. From this comparison, we find that GPT3.5 and finetuned GPT2-medium are comparable for planning the robot actions for each environment.

Our contributions are as follows: (1) We design the prompt templates to generate the task planning dataset for any domain. (2) We analyze the linguistic complexity of robotics datasets and a dataset for pretraining LLMs \cite{lima}, and discover a gap in complexity between the datasets. (3) By comparing LLMs with finetuned small LMs with our dataset, we find the potential of small LMs as the task planner.

\section{RELATED WORKS}
\subsection{Robotics with Large Language Models}
The rapid growth of LLMs is also having an impact on robotics. Recent research in scene understanding demonstrated that using LLMs outperforms using vision language models (VLMs) \cite{scene_understanding}. Furthermore, LLMs lead to higher accuracy in situational awareness tasks, by using text inputs to measure the ambiguity of user commands \cite{clara}. A few studies aim to design robots that can flexibly understand high-level user commands and infer current states through communication, by using LLMs \cite{roco, chatgpt_for_robot, tidybot}.

One of the most common applications of LLMs in robotics is task planning, which breaks down high-level commands into a sequence of low-level action steps. Some research enabled robots to generate plans in Pythonic code style using LLMs, given input user commands as predefined Python function library for robot actions \cite{chatgpt_for_robot, progprompt}. Also, there is research using LLMs as an intermediate model to ensure robust task planning \cite{tptu, sayplan}. Several other studies also employed LLMs for robot task planning purposes \cite{embodied_taskplan, task_motion_plan}.

Recent studies revealed that LLMs' performance is improved by prompting to reason in chains, known as Chain-of-Thought (CoT) \cite{cot_prompting}, and engineering the prompts. \cite{wt5, palm, zero_shot_reasoners, explanation}. Robot task planning shares similarities with CoT reasoning, where the reasoning process is broken down into intermediate steps to achieve a goal. Based on this, some researchers tried applying CoT to task planning. For instance, RT-2 \cite{rt2} employs large-scale VLMs to summarize the precise intent behind ambiguous user commands and generate sequences of action tokens. the Socratic model \cite{socratic}, SayCan \cite{saycan}, and PaLM-E \cite{palm_e} utilize LLMs to generate intermediate action steps based on abstract and ambiguous user commands. A common feature of these studies is that LLMs are used as a CoT reasoner to enable robots to plan to execute high-level commands in any domain.

However, do robots require such large-scale models capable of handling a wide range of NLP tasks? Existing studies often directly apply LLMs or LLM-based VLMs to robots, enabling them to perform high-level reasoning for commands from any environment. But in reality, modern robots still have limits in hardware to perform complex tasks in the real-world. While the ultimate goal is to execute complex tasks across various domains, it is equally important to consider the practical limitations of modern robots and choose appropriate LMs accordingly. Therefore, we narrow our focus to tasks that are within the current capabilities of robots and explore the possibility of smaller LMs.

\subsection{Language Models with Chain-of-Thought}
Recently, there has been a dramatic increase in the size of LMs, which we refer to as large-scale LMs (LLMs). It is demonstrated that LLMs performance is significantly enhanced by training LLMs to explain their reasoning process \cite{wt5}, and also engineering input prompts for LLMs to reason in chains \cite{palm, cot_prompting, zero_shot_reasoners, explanation}. These training and engineering methodologies are known as Chain-of-Thought (CoT) reasoning and CoT prompting, respectively. Researchers mainly focused on applying CoT prompting only to LLMs, since they assume that only large-scale LMs have the capabilities to describe intermediate reasoning steps. In practice, larger models absolutely outperform smaller models on CoT reasoning \cite{palm, cot_prompting, cot_good_for_bench, zero_shot_reasoners}, and from this, the spotlight has been on LLMs, resulting in less effort to efficiently improve the smaller LMs performance on CoT reasoning. 

Nonetheless, a few studies explored the potential of small LMs on CoT reasoning \cite{teacing_slm, slm_distillation_rationales, sci_cot, slm_specializing}. Magister et al. \cite{teacing_slm} explored the trade-off between model size and reasoning capabilities via knowledge distillation \cite{knowledge_distillation}, using LLMs as the teacher model and smaller LMs \cite{t5} as the student model. This study employed outputs from the teacher model as the training set to finetune the small LMs for CoT reasoning. The results showed that small LMs can learn CoT reasoning, as LLMs. To train small LMs to explain the reasoning process, Kang et al. \cite{slm_distillation_rationales} narrowed the models down to task-specific, and generated rationales for answers via knowledge distillation from LLMs. Ma et al. \cite{sci_cot} also tried to train small LMs to generate rationales via knowledge distillation from LLMs, by narrowing the domain down to scientific QA. Fu et al. \cite{slm_specializing} specialized LMs for a particular task and gave up some generalization capability of LMs.

Our approach shares similarities with these previous studies, the primary difference is that we focus on limiting both the task domain and the environment domain in which the robot can be deployed in practice. By training small LMs, such as GPT2-medium and GPT2-base \cite{gpt2}, whose parameters size are around 600 times smaller than GPT3.5 \cite{instruct_gpt}, for only a single task and environment, we anticipate that small LMs can be CoT reasoners, and also serve as task planners in robotics. Our goal is to develop small LMs capable of generating actionable steps for robots to understand and execute high-level commands within a `practical' environment where robots `actually' can be deployed, by limiting the domains that LMs train on. We aim to demonstrate the potential of small LMs in task planning.

\section{PROPOSED APPROACH}
\label{sec3:proposed_approach}

We build datasets to finetune small LMs for task planning in a single domain, such as kitchen and tabletop environments, via knowledge distillation from GPT3.5 \cite{instruct_gpt}, by CoT prompting on it. Our dataset consists of high-level commands and corresponding low-level actionable steps. To allow users to build a dataset for their domain, we propose prompt templates that can be adaptable to any environment and conditions. For our dataset, we also use the prompt templates with our environment and conditions. Our workflow is as follows:

\noindent \begin{itemize}[leftmargin=0.5cm]
\item [1.] \textbf{Domain-specific objects generation}
    \begin{enumerate}
        \item [(a)] Create a list of objects that can exist in a given domain (i.e., kitchen), by prompting LLMs.
    \end{enumerate}

\item [2.]\textbf{High-level command generation}
    \begin{enumerate}
        \item [(a)] Generate 20 high-level commands by prompting LLMs with the objects created in the previous step.
        \item [(b)] Repeat the process 2-(a) \textit{n} times until the desired number of commands is generated. Through this repeat, we get a total of $20*\textit{n}$ commands.
    \end{enumerate}

\item [3.]\textbf{Generating a sequence of action steps for each high-level command}
    \begin{enumerate}
        \item [(a)] Generate a sequence of actionable steps to execute the high-level command, by CoT prompting LLMs.
        \item [(b)] Repeat the process 3-(a) $20*\textit{n}$ times, until all high-level commands have a corresponding action steps.
    \end{enumerate}

\item [4.]\textbf{Fine-tuning small language models}
    \begin{enumerate}
        \item [(a)] Finetune the small LMs to generate intermediate actionable steps for the high-level commands, using the command-steps pairs dataset.
    \end{enumerate}
\end{itemize}

\begin{figure}[t]
    \centering
    \includegraphics[width=0.95\columnwidth]{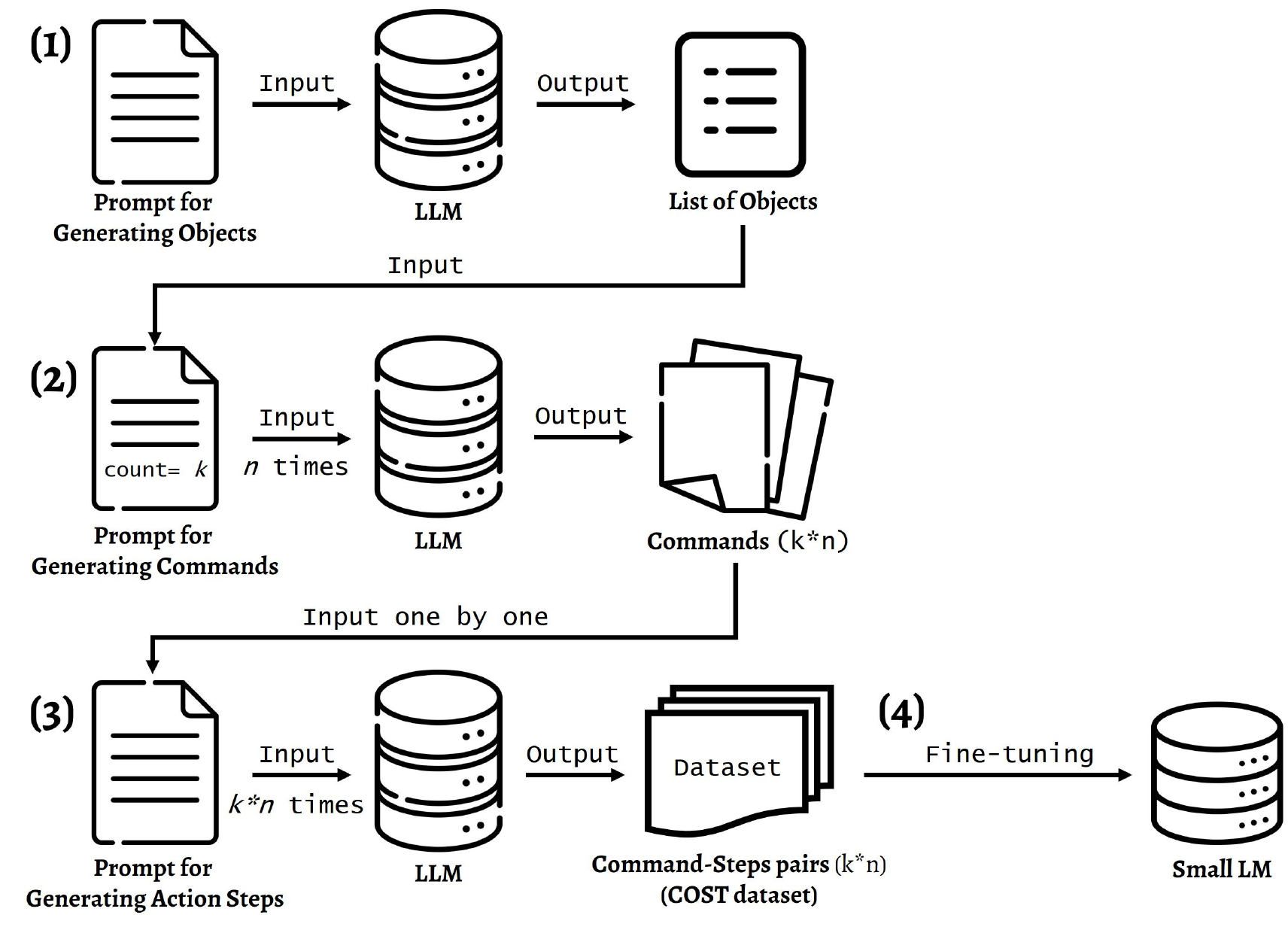}
    \vspace*{-0.3cm}
    \caption{A flow chart describing how our COmmand-STep dataset (COST) is generated. Detailed description can be found in Sec. \ref{sec3:proposed_approach}.}
    \label{fig:dataset_flow}

\end{figure}

The overall generation process of our dataset is summarized in Figure \ref{fig:dataset_flow}. We name our dataset as COST, COmmand-STeps dataset.

\subsection{Objects List Generation}
We designed a prompt template for generating objects that can exist in a given domain. Our prompt template allows the user to generate a list of objects in whatever domain they plan to deploy the robot (i.e., Kitchen). If the user already has the list of objects for their own domain, one can skip this step.

To create objects suitable for robot manipulation, we describe the hardware specification of the robot in the prompt. To guide LLMs in generating appropriate objects, we also instruct LLMs to briefly describe the use of each object. Additionally, we give a few output samples in the prompts to represent what we desire as output.

Figure \ref{fig1:objs_prompt} shows our prompt template for object generation. In this prompt, engineering [Condition of the object] and the \{sample objects\} is crucial. For instance, if the user fills the prompt template for the kitchen without any [Condition of the object], and just gives `knife' as {A sample object}, LLMs focus only on kitchen tools. However, if the user provides conditions such as `the object can be a cooking ingredient like a carrot.', and gives balanced output samples, `potato' and `knife', LLMs generate tools or ingredients as objects, in balance. What matters is that the users can clearly describe the object features they want to generate for their prompt.

\definecolor{shallowgreen}{RGB}{22, 173, 30}
\begin{figure}[t]
  \centering
  \framebox[3.3in]{\parbox{3.15in}{
\begin{scriptsize}
\begin{spacing}{1}
\vspace{0.05cm}
List up  \textcolor{shallowgreen}{\{count\}} objects that the robot can PICK UP, GRAB, and use in the “\textcolor{shallowgreen}{\{domain\}}". For each object, describe why that object could be used by the robot in that place.\\

\vspace{-3.5pt}
[Situation]\\
\textcolor{shallowgreen}{\# There is the robot in the \{domain\}.}\\

\vspace{-3.5pt}
[Condition of the object]\\
\textcolor{shallowgreen}{\# Describe the conditions of objects you want to generate.}\\

\vspace{-3.5pt}
[Information about the robot]\\
\textcolor{shallowgreen}{\# Describe briefly the hardware specification of the robot.}\\

\vspace{-3.5pt}
[Outputs]\\
1. \textcolor{shallowgreen}{\{sample objects\}}\\
: \textcolor{shallowgreen}{\{A sample description of how the object could be used\}}\\

2.\\
\vspace{-3.5pt}
\end{spacing}
\end{scriptsize}
}}

\caption{The prompt template to generate an object list in the given domain. The green lines are where the user needs to fill in. The prompt requires the user to specify their domain, the number of objects to be generated, the conditions of the objects, information about the robot, and output samples.}
    
\label{fig1:objs_prompt}
\end{figure}

\definecolor{shallowgreen}{RGB}{22, 173, 30}
\begin{figure}[t!]
  \centering
  \framebox[3.3in]{\parbox{3.15in}{

\begin{scriptsize} 
\vspace{0.05cm}
\begin{spacing}{1}
I want to collect examples of instructions that a person might give to a robot in \textcolor{shallowgreen}{\{domain\}}. You need to generate \textcolor{shallowgreen}{\{count\}} sets of instructions using the [Object list] provided. Each set should include $<$USED\_OBJECTS$>$ and $<$INSTRUCTION$>$. Please adhere to the guidelines mentioned in [Condition of the instructions generated] below.\\

\vspace{-3.5pt}
I've given you an [Output template] to follow. You should replace the placeholders $<$USED\_OBJECTS$>$ and $<$INSTRUCTION$>$, using the [Example of output] as a reference. \\

\vspace{-3.5pt}
[Conditions of the commands to be generated]\\
* Each $<$INSTRUCTION$>$ should be a concise single sentence.\\
* The instructions should be constructed by following these steps:\\
Step 1. Select \textit{n} or more objects to use from the [Object list].\\
Step 2. Create an instruction using the selected objects from step 1.\\

\vspace{-3.5pt}
[Example of output]\\
(\{Sample objects used in the command\}) \{A sample command\}\\

\vspace{-3.5pt}
[Output template]\\
($<$USED\_OBJECTS$>$), $<$INSTRUCTION$>$\\

\vspace{-3.5pt}
[Note]\\
\textcolor{shallowgreen}{\# Describe notification which you’d like LLMs to consider when generating commands.}\\

\vspace{-3.5pt}
[Object list]\\
\textcolor{shallowgreen}{\# Input your objects list in here.} \\

\vspace{-3.5pt}
--------------------------------

[Output]\\
1.\\
\vspace{-3.5pt}
\end{spacing}
\end{scriptsize}
}}

    \caption{The prompt template to generate high-level commands using the object list. We first describe the task that LLM should do, and present conditions for generating, output examples, an output template and give an object list as an input. The green lines are where the user needs to fill in.}
    \label{fig:inst_prompt}
\end{figure}
\definecolor{shallowblue}{RGB}{35, 102, 226}
\begin{figure}[t]
  \centering
  \framebox[3.3in]{\parbox{3.15in}{
\begin{scriptsize} 
\begin{spacing}{1}
\vspace{0.05cm}
\textcolor{shallowblue}{You have been given "Objects in front of the robot" and "Command". Your task is to generate "Action Steps".}\\

\vspace{-3.5pt}
An [Output template] will be provided for your output, and $<$MASK$>$ is a placeholder for content. Please see the \textcolor{shallowblue}{[Information about the robot] and} [Note] section for additional guidance.\\

\vspace{-3.5pt}
\textcolor{shallowblue}{[Information about the robot]} \\
\textcolor{shallowgreen}{\# Describe physical capability of your robot.}\\

\vspace{-3.5pt}
[Output template]\\
Action Steps= $<$MASK$>$\\

\vspace{-3.5pt}
[Example of Input and Output]\\
*Input:\\
Command= \{A sample command\}\\
\textcolor{shallowblue}{Objects in front of the robot=\{A sample object list\}}\\

\vspace{-3.5pt}
*Output:\\
Action Steps=\\
Step 1. \{A sample step 1\} (ACTION : \{Action in step1\}, TARGET: \{Target in step 1 action\})\\
…\\
Step \textit{n}. \{A sample step \textit{n}\} (ACTION : \{Action in step \textit{n}\}, TARGET: \{Target in step \textit{n} action\})\\

\vspace{-3.5pt}
[Note]\\
\textcolor{shallowgreen}{\# Describe conditions which you’d like LLMs to consider when generating steps.}\\

\vspace{-3.5pt}
--------------------------------

*Input:\\
Command= \textcolor{shallowgreen}{\{A input command\}}\\
\textcolor{shallowblue}{Objects in front of the robot= \textcolor{shallowgreen}{\{A input objects list\}}}\\

\vspace{-3.5pt}
*Output:\\

\vspace{-3.5pt}
\end{spacing}
\end{scriptsize}
}}
    \caption{The prompt template for generating action steps for each high-level command. The blue lines are only for the prompt where the available objects are fixed, and the black lines are common to prompts in Fig \ref{fig5:steps_prompt_tabletop} and \ref{fig6:steps_prompt_kitchen}. The green lines are where the user needs to fill in.}
    \label{fig5:steps_prompt_tabletop}
\end{figure}

\definecolor{shalloworange}{RGB}{249, 129, 19}
\begin{figure}[t]
  \centering
  \framebox[3.3in]{\parbox{3.15in}{
\begin{scriptsize} 
\begin{spacing}{1}
\vspace{0.05cm}
\textcolor{shalloworange}{You have been given a "Command" by one person to a robot. Your task is to generate two outputs: 1. "Action Steps" 2. "Required Objects”}\\

\vspace{-3.5pt}
An [Output template] will be provided for your output, and $<$MASK$>$ is a placeholder for content. Please see the [Note] section for additional guidance.\\

\vspace{-3.5pt}
[Output template]\\
Action Steps= $<$MASK$>$\\
\textcolor{shalloworange}{Required Objects= $<$MASK$>$}\\

\vspace{-3.5pt}
[Example of Input and Output]\\
*Input:\\
Command= \{A sample command\}\\

\vspace{-3.5pt}
*Output:\\
Action Steps=\\
Step 1. \{A sample step 1\} (ACTION : \{Action in step1\}, TARGET: \{Target in step 1 action\})\\
…\\
Step \textit{n}. \{A sample step \textit{n}\} (ACTION : \{Action in step \textit{n}\}, TARGET: \{Target in step \textit{n} action\})\\
\textcolor{shalloworange}{Required Objects= \{A sample list of required objects\}}\\

\vspace{-3.5pt}
[Note]\\
\textcolor{shallowgreen}{\# Describe conditions which you’d like LLMs to consider when generating steps.}\\

\vspace{-3.5pt}
--------------------------------

*Input:\\
Command= \textcolor{shallowgreen}{\{A input command\}}\\

\vspace{-3.5pt}
*Output:\\

\vspace{-3.5pt}
\end{spacing}
\end{scriptsize}
}}
    \caption{The prompt template for generating action steps for each high-level command. The blue lines are only for the prompt where the available objects are not fixed, and the black lines are common to prompts in Fig \ref{fig5:steps_prompt_tabletop} and \ref{fig6:steps_prompt_kitchen}. The green lines are where the user needs to fill in.}
    \label{fig6:steps_prompt_kitchen}
\end{figure}
\subsection{High-level Commands Generation}
`High-level command' refers to the abstract command that includes a lot of implicit contents and, in this study, also refers to the command that requires multiple action steps to achieve the goal. We generate more than 20,000 high-level commands for our COST datasets. Generating all commands by inputting a single prompt is challenging due to the limitation of LLMs' maximum output length and the tendency of noise to increase towards the latter part of the output. Therefore, we generate only 20 high-level commands per a single prompt input, and repeatedly call the GPT3.5 API until the desired number of commands are generated.

Figure \ref{fig:inst_prompt} shows our prompt template to generate high-level commands. To fix the output format, we provide the output template and instruct LLMs to fill in the placeholders in the template, by referring to the guidelines presented in \cite{prompt_engineering}. To guide LLMs to use only the objects in [Object list], we give [Conditions of the commands to be generated] as a guideline in the prompt, which suggests the following generation order: First, select objects from [Object list], then, generate a high-level command only using the selected objects. If we do not give the guideline, LLMs will often create new objects that are not included in [Object list] when generating commands. However, it is challenging to eliminate the possibility of new objects creation, because LLMs are probability models. If the user desires to avoid generating commands using objects outside the given list, a post-processing will be required. As we fixed the output format through prompting, a post-processing would be relatively straightforward. However, we rather recommend using new objects that LLMs generate by breaking the rules in the prompt. In our experiment, we keep adding new objects to the [Object List] to be used in the next prompt input. This guarantees us to generate more varied high-level commands from LLMs, but the number of objects in the list keeps increasing. So, we randomly sample \textit{N} objects from the entire object list as the [Object list] for each prompt input.

The users can control `How abstractly generate the commands' via [Example of output]. If the user wants highly abstract commands, it is recommended to avoid clarifying the name of used objects in \{A sample command\}, such as `Make a sandwich.'.  On the other hand, if the user prefers relatively explicit commands, it is recommended to clarify the name of used objects, such as `Make a sandwich with bread, tomato, and cheese, after cutting the tomato.'.

\subsection{Action Steps Generation}
Figure \ref{fig5:steps_prompt_tabletop} and \ref{fig6:steps_prompt_kitchen} are our prompt templates to generate `Action steps' for each high-level command. Action steps are defined as the process of breaking down high-level commands into intuitive and simple actionable steps for the robot, to achieve the goal of the command \cite{saycan}. Based on this definition, we can consider action steps as CoT reasoning, in that it is a process of breaking down high-level instruction into intermediate steps to achieve the answer (or goal). Each action step consists of a short description of what the robot needs to do at this stage, along with `ACTION' which the robot should perform, and `TARGET' which the robot should aim. Specifying ACTION and TARGET helps avoid noise, such as a step with more than one action.

By prompting, we instruct the LLMs to generate a sequence of actionable steps required for the robot to execute a high-level command. We design the prompts to input only one high-level command at a time, and to output a list of action steps associated with the input command. Since we should generate action steps lists for every high-level command, we repeatedly called GPT3.5 API until we got an action steps list for every command.

We adopt guidelines \cite{prompt_engineering} that fill in placeholders to standardize the output format. We design two versions of the prompt to cover different cases: When the objects to be used in action steps are \textbf{fixed} or \textbf{flexible}. The former prompt, Figure \ref{fig5:steps_prompt_tabletop}, is designed to generate action steps using only given objects, and the latter prompt, Figure \ref{fig6:steps_prompt_kitchen}, is designed to generate not only the action steps but also the objects required by the action steps, for the input command.

We use the prompt in Figure \ref{fig5:steps_prompt_tabletop} to build the COST dataset for a tabletop domain. For general real-world environments, such as kitchens, we use the prompt in Figure \ref{fig6:steps_prompt_kitchen}. In tabletop, robot actions are limited to pick-and-place, so we provide the [Information about the robot] section only for the prompt in Figure \ref{fig5:steps_prompt_tabletop} to describe the physical capabilities of the robot, to ensure that LLMs generate Action steps consisting of actions that the robot can perform in practice.

The crucial parts that are common to both versions of the prompt are [Example of input and output] and [Note]. For instance, to ensure action steps in a tabletop environment consist of only pick-and-place actions and to avoid consecutive picking actions, the user should design an example that reflects all of these conditions. It is also crucial to mention in [Note] the conditions that LLMs should follow when generating action steps. For instance, if one needs to generate action steps by limiting the number of objects the robot can hold at once, this rule can be described in [Note].

\subsection{Fine-tuning Small Language Models}
We build our COST datasets for the tabletop and kitchen environments, by prompting LLMs using prompt templates Fig \ref{fig:inst_prompt}-\ref{fig6:steps_prompt_kitchen}. We generate datasets for just two domains, but note that our prompt templates allows anyone to generate a dataset for their task domain. Using the dataset, we fine-tune LMs to generate action steps. Then we verified if small LMs can plan to execute the high-level command, when the environment and task domain are limited.

\section{DATASET}
\subsection{Comparison of Datasets for Robotics and LLMs}
We analyze the complexity between the text datasets in robotics and the dataset used for pretraining LLMs. Based on the result of the analysis, we aim to verify whether the large-scale LMs with more than 100 billion parameters were indeed essential for processing the commands by the robot.

We use Lima \cite{lima} as a representative of the dataset for pretraining LLMs, and use RT-1 \cite{rt1}, RoboVQA \cite{robovqa}, BridgeData V1 and V2 \cite{bridge1, bridge2}, and the Language Table dataset \cite{lt_dataset} for simulator as the robotics datasets. We only utilized text data from these robotics datasets for our comparison. 

To compare the readability levels among datasets, we employed the Flesch-Kincaid grade level formula as a metric. The Flesch-Kincaid test is the metric used to determine what school grade is the document appropriate to read. The metric score is calculated based on the number of words per sentence and the number of syllables per word. The result of 1 to 18 means the document is suitable for reading in grades 1st to 18th. \textbf{Kincaid} in Table \ref{table:complexity} shows that while robotics datasets are suitable for reading at 1st to 3rd grade, Lima requires a much higher readability level. To measure the complexity per sentence, we calculated the depth of the syntactic tree. The syntactic tree is a tree diagram that depicts the syntactic structure of a sentence. If the depth of the syntactic tree is large, it means that the sentence consists of many phrases and clauses. In Table \ref{table:complexity}, \textbf{Syntactic tree} denotes the average depth of the syntactic tree for all sentences in the dataset, with the numbers in parentheses indicating variance. It shows that the sentences in Lima are structurally deeper than all other robotics datasets. Lima also has the largest depth's variance than the robotics dataset, and it can be interpreted as Lima consisting of the most diverse sentence structures. We also analyzed the vocabulary size by counting the number of unique tokens (types) in each dataset. \textbf{\#Types} in Table \ref{table:complexity} shows that Lima contains a larger vocabulary than the robotics datasets.

\begin{table}[t!]
\centering
\small
    \begin{tabular}{l|ccc}
    \hline
        \multirow{2}{*}{\textbf{Datasets}}
        & \multicolumn{3}{c}{\textbf{Complexity}} \\
        \cline{2-4}
        & \textbf{Kincaid} & \textbf{Syntactic tree} & \textbf{\#Types} \\

        \hline
        {Lima}            &{7.91} &{5.56\small{(4.4)}} &{29962} \\
        \hline
        {RT1}             &{1.89} &{4.3\small{(0.82)}} &{51}  \\
        {RoboVQA}         &{1.86} &{4.44\small{(1.58)}} &{2624}  \\
        {Bridge}          &{2.7} &{5.36\small{(1.42)}} &{2678}  \\
        {LT\small{(sim)}} &{3.54} &{5.28\small{(1.24)}} &{843} \\
        
        \hline
        
        \cellcolor{black!13}{Kitchen, Command} &\cellcolor{black!5}{3.07} &\cellcolor{black!5}{4.6\small{(0.8)}} &\cellcolor{black!5}{2239}  \\
        
        \cellcolor{black!13}{Kitchen, Steps}  &\cellcolor{black!5}{1.32} &\cellcolor{black!5}{3.84\small{(0.71)}} &\cellcolor{black!5}{2979}  \\
        
        \cellcolor{black!13}{Tabletop, Command} &\cellcolor{black!5}{4.81} &\cellcolor{black!5}{5.71\small{(1.94)}} &\cellcolor{black!5}{1410}  \\
        
        \cellcolor{black!13}{Tabletop, Steps} &\cellcolor{black!5}{1.79} &\cellcolor{black!5}{4.45\small{(2.09)}} &\cellcolor{black!5}{990}  \\
        \hline
    
    \end{tabular}
\caption{Dataset Complexity Comparison. The gray cell is for our datasets. For more details on Lima and robotics datasets, see Sec. 4.1, and for more details on our datasets, see Sec. 4.2.}
\label{table:complexity}
\vspace{-0.5cm}
\end{table}

\begin{table}[t!]
\small
\centering
    \begin{tabular}{l|ccc}
    \hline
        \multirow{2}{*}{\textbf{Datasets}}
        & \multicolumn{3}{c}{\textbf{Part-of-Speech}} \\
        \cline{2-4}
        & \textbf{NN}
        & \textbf{VB}
        & \textbf{AD} \\

        \hline
        {Lima}            &{10.59} &{8.08} &{5.34} \\
        \hline
        
        {RT1}             &{3.08} &{0.85} &{2.33}  \\
        {RoboVQA}         &{3.63} &{2.37} &{2.08}  \\
        {Bridge}          &{3.58} &{3.3} &{1.95}  \\
        {LT\small{(sim)}} &{2.3} &{1.57} &{0.75} \\
        \hline
        
        \cellcolor{black!13}{Kitchen, Command} &\cellcolor{black!5}{5.34} &\cellcolor{black!5}{4.79} &\cellcolor{black!5}{3.54}  \\
        \cellcolor{black!13}{Tabletop, Command} &\cellcolor{black!5}{3.96} &\cellcolor{black!5}{3.29} &\cellcolor{black!5}{2.49}  \\

        \hline
    \end{tabular}
\caption{Average entropy per PoS. The gray cell is for our datasets. For more details on the Lima and robotics datasets, see Sec. 4.1, and for more details on our datasets, see Sec. 4.2.}
\label{table:pos_entropy}
\vspace{-0.3cm}
\end{table}

We examine the bias in vocabulary usage in each part of speech (PoS) for each dataset. To analyze it, we calculated the entropy for the occurrence frequency of words per PoS. We interpret that higher entropy means that diverse words are evenly used, and lower entropy means that only a few specific words are frequently used. The formula for calculating the entropy $H(\textit{W})$ is as follows:

\vspace*{-12pt}
\begin{gather}
H(\textit{W})= -\sum_{\textit{w} \in \textit{W}}p(\textit{w})\cdot log p(\textit{w})\\
p(\textit{w})= \frac{\text{Occurrence number of \textit{w}}}{\text{Total number of tokens}}
\end{gather}
\vspace*{-8pt}

where $\textit{W}$ is a set of total words (types) in each PoS in the document (dataset), \textit{w} is a word in $\textit{W}$, and $p(\textit{w})$ is occurrence probability of \textit{w} in certain PoS in the document.

Table \ref{table:pos_entropy} shows average entropy per PoS. \textit{NN}, \textit{VB}, and  \textit{AD} mean a set of noun families, a set of verb families, and a set of adjective and adverb families, respectively. Across all PoS, the entropy for the robotics dataset is significantly lower compared to Lima, and we analyze it because the robotics datasets consist mostly of commands that can be executed with limited actions and objects in a given domain. The results show that robotics datasets utilize only a few specific words noticeably more often.

Table \ref{table:complexity} and \ref{table:pos_entropy} show that the dataset for pretraining LLMs is more challenging to read, consists of diverse and complex sentences, and requires more vocabulary, compared to robotics datasets. Through the results, we can identify a gap in complexity between robotics datasets and datasets for pretraining LLMs. Based on this, we expect that smaller LMs also be able to comprehend input commands and plan effectively for the tasks, if it can be demonstrated that a small LMs can learn CoT reasoning for robotics planning within a single domain.

\subsection{Our Dataset Generated by LLMs}
 \begin{figure}[t!]
  \centering
  \framebox[3.3in]{\parbox{3.15in}{
  \begin{scriptsize}
  \begin{spacing}{1}
\vspace{0.1cm}
\textbf{Objects=} ['Bowl', 'Whisk', 'Table', 'Egg', 'Apple'] \\
\textbf{Command=} Beat the egg in a bowl. \\
\textbf{Action Step= }\\
Step 1. PICK up the egg. (ACTION: Pick $\mid$ TARGET: Egg) \\
Step 2. CRACK the egg into the bowl. (ACTION: Crack $\mid$ TARGET: Bowl) \\
Step 3. PICK up the whisk. (ACTION: Pick $\mid$ TARGET: Whisk) \\
Step 4. BEAT the egg in the bowl using the whisk. (ACTION: Beat $\mid$ TARGET: Whisk, Egg, Bowl)
\vspace{0.1cm}
\end{spacing}
\end{scriptsize}
}}
\caption{A sample of our COST dataset for kitchen domain. For a detailed description of the sample, please refer to Sec. \ref{sec4:our_dataset}}
\label{fig:sample_data}
\end{figure}
\label{sec4:our_dataset}

We find that the commands in robotics are simpler than what LLMs learn, from the result, we expect that small LMs can learn well for task planning. To finetune the planning process on small LMs, we need our command-steps datasets of similar complexity to datasets for robotics. So, we build the COST dataset using GPT3.5 \cite{instruct_gpt}. Please refer to Sec. \ref{sec3:proposed_approach}. for the detailed process of building this dataset. 

The gray cells in Table \ref{table:complexity} and Table \ref{table:pos_entropy} indicate the complexity of our dataset. `Kitchen, Command' and `Tabletop, Command' refer to the high-level command set for each environment, and `Kitchen, Steps' and `Tabletop, Steps' refer to the steps set for each environment, respectively. Table \ref{table:complexity} shows that the complexity of our command set is similar to that of robotics datasets. However, looking at the \#Types in table \ref{table:complexity} and the entropy per PoS in Table \ref{table:pos_entropy}, we can see that our command set has a larger and more diverse vocabulary than robotics datasets. Considering that this vocabulary is all for one domain, we can say that our datasets contain richer vocabulary for a single domain than the robotics datasets. We can also see that the steps set contains easier sentences than our command set and robotics datasets.

Figure \ref{fig:sample_data} depicts sample data in the kitchen domain. \textbf{Objects} is an objects list in front of the robot. In order to train LMs to select only the `necessary' items from a given objects, Objects contains not only the necessary items to execute the command, but also unnecessary items. Objects are followed by \textbf{Command} and then \textbf{Action Steps}. Command is an input instruction to the robots, and Action Steps is a set of intermediate steps to execute the command.

\section{EXPERIMENTS}
\subsection{Example Results}
\begin{figure}[t!]
  \centering
  \begin{scriptsize}
  \begin{spacing}{1}
  \framebox[3.3in]{\parbox{3.15in}{
\vspace{0.1cm}
\textbf{Input: } \\
Objects= ['Yellow semicircle block', 'Red circle block', 'Red semicircle block', 'Yellow bowl', 'Red bowl']\\
Command= Arrange semicircle blocks in the same colored bowl as the picked block.\\
Action steps= \\

\vspace{-3.5pt}
-------------------------------\\
\textbf{Output: }\\
Step 1. PICK up the red semicircle block. (ACTION: Pick $|$ TARGET: Red semicircle block)\\
Step 2. PLACE the red semicircle block in the red bowl. (ACTION: Place $|$ TARGET: Red bowl)\\
Step 3. PICK up the yellow semicircle block. (ACTION: Pick $|$ TARGET: Yellow semicircle block)\\
Step 4. PLACE the yellow semicircle block in the yellow bowl. (ACTION: Place $|$ TARGET: Yellow bowl)
\vspace{0.1cm}}}
\end{spacing}
\end{scriptsize}
\caption{An example result in tabletop domain. When the \textbf{Input} is fed into the model, it predicts what comes after the Input and return the \textbf{Output}.}
\label{fig:tabletop_output_sample}
\end{figure}

\begin{figure}[t!]
    \centering
\begin{scriptsize}
\begin{spacing}{1}
    \framebox[3.3in]{\parbox{3.15in}{
\vspace{0.1cm}
\textbf{Input: } \\
Objects= ['Cucumber', 'Lettuce', 'Tray', 'Sink', 'Knife', 'Cup']\\
Command= Slice the washed lettuce and place it on tray.\\
Action steps= \\

\vspace{-3.5pt}
-------------------------------\\
\textbf{Output: }\\
Step 1. PICK up the lettuce. (ACTION: Pick $|$ TARGET: Lettuce)\\
Step 2. MOVE to the sink. (ACTION: Move $|$ TARGET: Sink)\\
Step 3. WASH the lettuce under running water. (ACTION: Wash $|$ TARGET: Lettuce)\\
Step 4. PICK up the knife. (ACTION: Pick $|$ TARGET: Knife)\\
Step 5. SLICE the lettuce using the knife. (ACTION: Slice $|$ TARGET: Knife, Lettuce)\\
Step 6. PLACE the sliced lettuce on the tray. (ACTION: Place $|$ TARGET: Sliced lettuce, On the tray)
\vspace{0.1cm}  }}
\end{spacing}
\end{scriptsize}
\caption{An example result in kitchen domain. When the \textbf{Input} is fed into the model, it predicts what comes after the Input and return the \textbf{Output}.}
\label{fig:kitchen_output_sample}
\end{figure}
\begin{table}[t]
\centering
\small
    \begin{tabular}{l|c}
    \hline
    {\textbf{Models}} & {\textbf{Success Rate (\%)}} \\
    \hline
    \hline
        GPT2-base &{50}\\
        GPT2-medium &{85}\\
        GPT3.5 &{70 (90)} \\
        GPT4 &{95 (100)}\\
        
        \hline
    \end{tabular}
\caption{The success rate for 20 test commands for the tabletop domain. For GPT3.5 and GPT4, The numbers denote the result from an prompt from \cite{socratic}, and from our prompt, respectively.}
\label{table:tabletop_success_rate}
\vspace{-0.35cm}
\end{table}

Figure \ref{fig:tabletop_output_sample} and \ref{fig:kitchen_output_sample} show examples of results from finetuned GPT2-medium with our dataset for the tabletop and kitchen domain, respectively. The inputs are an unseen command and the random objects including items required to execute the command. 

To evaluate small LMs finetuned with our tabletop dataset, we aim to apply them on the tabletop simulator. The possible actions in the tabletop simulator are only pick-and-place, so, we restrict all actions in the tabletop dataset to pick or place. For the kitchen dataset, we constrain the possible actions to pick, place, move, and utilize the objects.

\subsection{Results in the Tabletop Simulator}
\begin{figure*}[t]
\centering
\includegraphics[width=\textwidth]{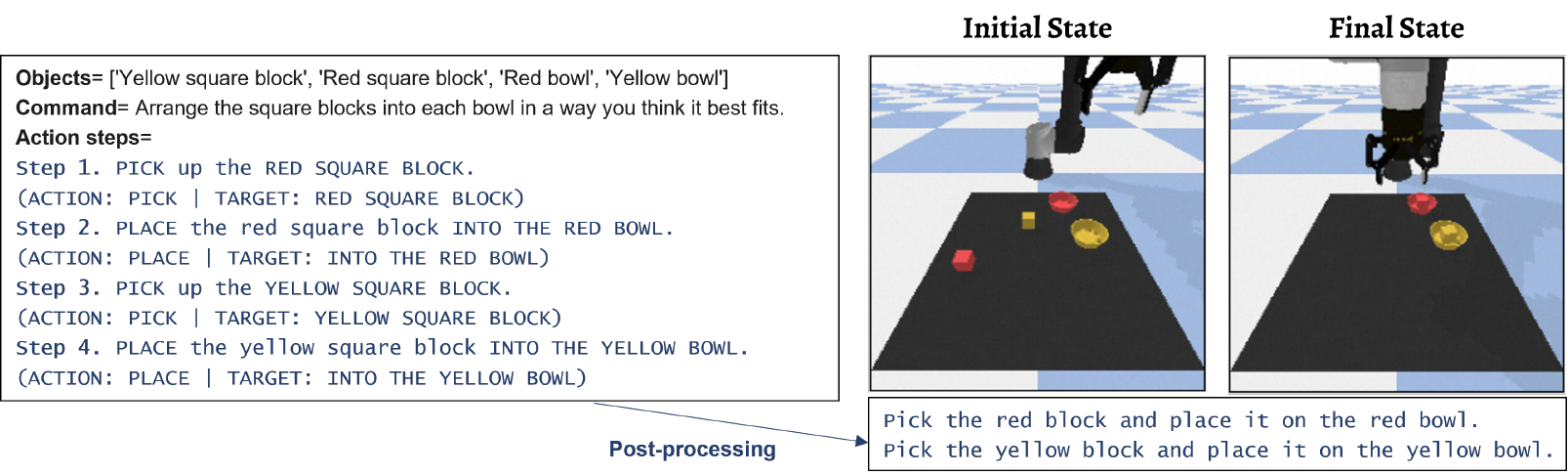}
\vspace*{-0.5cm}
\caption{The sample result of finetuned GPT2-medium in the tabletop simulator. In the first text box, the black text is the input to the model, and the blue text is the output of the model. We post-process the output to input it in CLIPort \cite{cliport}. The `Initial State' image refers to the start scene on the simulator, and the `Final State' image refers to the end scene after performing all \textbf{Action steps}.}
\label{fig:sample_img_tabletop_experiment}
\end{figure*}
\label{sec:result_in_Tabletop}
We aim to compare the results of small LMs finetuned on our tabletop dataset, with the results of Socratic models (SMs) \cite{socratic}, which plan in chain for the command using LLMs. For this, we finetune GPT2-base and GPT2-medium \cite{gpt2} on our tabletop dataset, and experiment on the tabletop simulator with the finetuned LMs outputs and the LLMs outputs. To apply generated \textbf{Action steps} to the tabletop simulator, it needs to go through CLIPort, so we need to post-process them to fit CLIPort input format. Figure \ref{fig:sample_img_tabletop_experiment} shows how we post-process them.

We evaluate LMs for 20 test data, which consist of unseen commands and random objects. The order of the experiments is as follows: (1) Randomly spawn objects on the table in the simulator. (2) Input a command to the LMs, and get a sequence of intermediate actionable steps. (3) Post-process the output to fit in CLIPort input format, and enter it into CLIPort. (4) The robot performs the action steps based on the output of CLIPort.

We compare the initial and final scenes after performing all action steps, and manually judged if the robot successfully executed the input commands. We used the GPT3.5 and GPT4 outputs on the prompts in \cite{socratic} as the SMs results, and also verified the outputs from our prompt in Figure \ref{fig5:steps_prompt_tabletop}. 

Table \ref{table:tabletop_success_rate} presents the success rate per model for the 20 test sets. GPT2-base has a noticeably lower success rate, and GPT4 has a noticeably higher success rate, but the notable point is that the GPT2-medium achieves a higher success rate than GPT3.5 on SMs. However, GPT3.5 on our prompt achieves a higher success rate than GPT2-medium. Consequently, despite the large gap in the number of parameters, GPT2-medium and GPT3.5 are comparable in tabletop environments, and GPT3.5 has a bit of unstable performance because it depends on prompting.

\subsection{User Study}
\begin{figure}[t!]
    \centering
    \includegraphics[width=0.95\columnwidth]{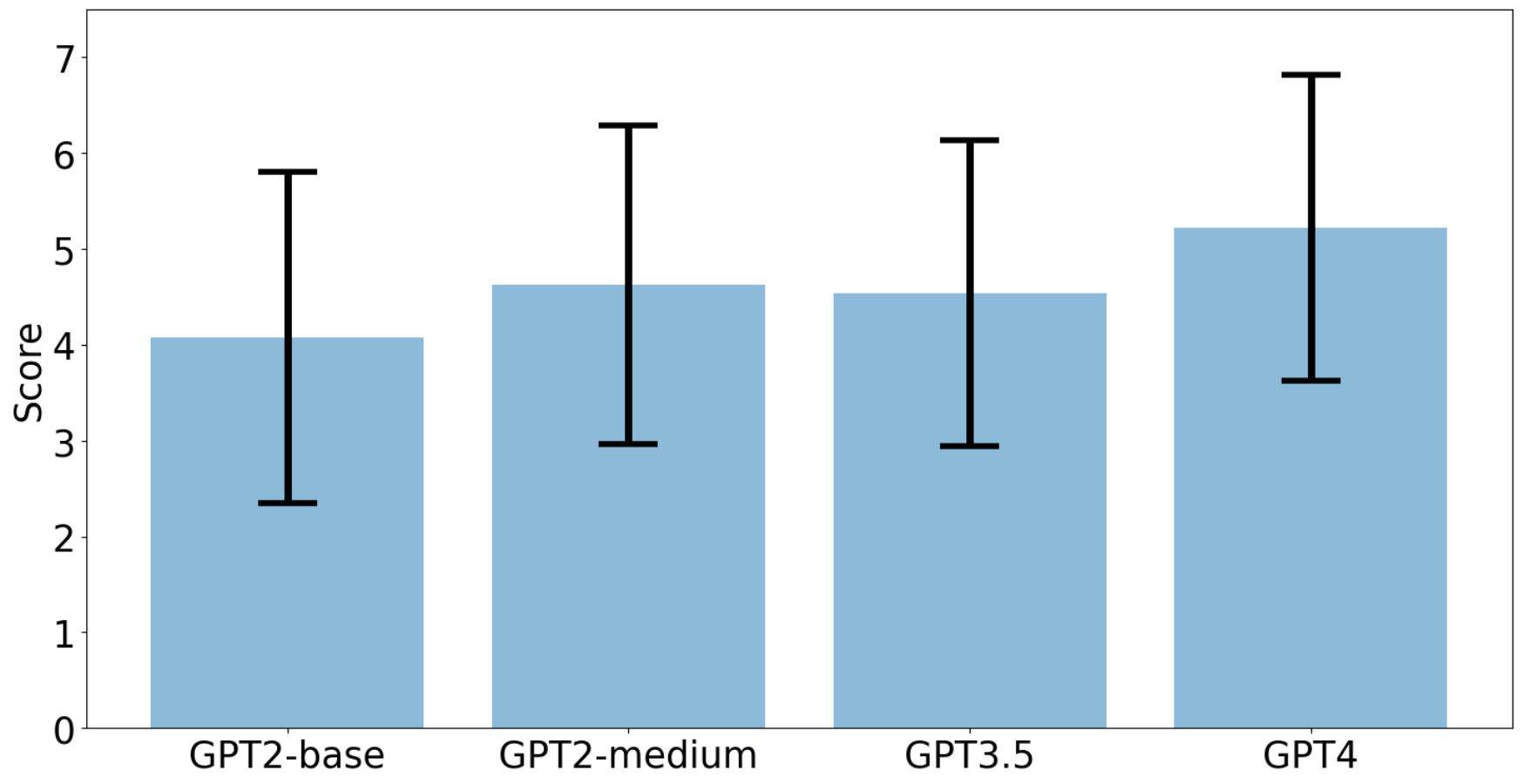}
    
    \caption{Bar graph for survey results. The y-axis represents a score scale from 1(Very poor) to 7(Excellent).}
    \label{fig:barplot_userstudy}
\end{figure}

To evaluate whether small LMs finetuned on our datasets can plan effectively for high-level commands as LLMs, we conducted a survey. The survey includes input-outputs samples for the kitchen domain. We employ GPT2-base and GPT2-medium as small LMs, and employ GPT3.5 and GPT4 as LLMs. To get the outputs under the same conditions as in small LMs, we use the LLMs' output returned from inputting the prompt \ref{fig:inst_prompt}. Participants receive the input objects, input commands, and each model's outputs for the input, without being informed of which model each output was from, and asked to rate each output on a scale of 1 to 7. The scoring criteria are as follows: First, whether the output steps are appropriate to execute the input command. Second, the output steps are represented in small enough steps for the robot to follow. Third, the output steps can be performed using the appropriate things in given objects. The survey consisted of 3 different input sets and their outputs. We built three versions of the survey with different sample sets and surveyed 40 people per survey. The results of the three surveys are combined to get the overall results. 

Figure \ref{fig:barplot_userstudy} is the bar graph for the survey result. It shows that GPT2-base has a markedly lower score, and GPT4 has a markedly higher score. However, interestingly, there is little difference in scores between GPT2-medium and GPT3.5, and even GPT2-medium scored more consistently higher. The \textit{p}-value between GPT2-medium and GPT3.5 is around 0.42, which means that there is no statistical difference between the results of the two models. Consequently, in a given domain, GPT2-medium fine-tuned for planning can reason in chains with nearly equivalent performance to GPT3.5.

\section{CONCLUSIONS}
In this work, we suggest the question for the discrepancy between the low complexity of commands in robotics and the advanced capability of LLMs, and explore the potential for task planning on small LMs in a single domain. To validate the question, first, we build the COmmand-STeps dataset (COST) on a single domain and finetuned small LMs with it. Then, we experiment to compare the finetuned small LMs with LLMs in not only tabletop simulator but also user study for the kitchen environment. As a result, we find that domain-specific GPT2-medium and domain-agnostic GPT3.5 are comparable for planning the robot actions in the given environment. We argue that LLMs may not always be the best choice for task planning in robotics, and we also hope that researchers will positively consider using small LMs to fit the environment in which they deploy their robots. 

\addtolength{\textheight}{-12cm}   

\bibliographystyle{IEEEtran}
\bibliography{citation}

\end{document}